\documentclass[11pt,a4paper]{article}
\usepackage[hyperref]{acl2021}
\usepackage{times}
\usepackage{latexsym}

\usepackage{microtype}

\aclfinalcopy 

\usepackage{booktabs}
\usepackage{amsmath}
\usepackage{amssymb}

\usepackage{caption}
\usepackage{graphicx}
\newcommand{\abs}[1]{\left \lvert #1 \right \rvert}

\title{Why Can You Lay Off Heads? \\ Investigating How BERT Heads Transfer}

\author{Ting-Rui Chiang \quad Yun-Nung Chen \\ 
 Department of Computer Science and Information Engineering \\
 National Taiwan University, Taipei, Taiwan \\
 \texttt{r07922052@csie.ntu.edu.tw} \quad \texttt{y.v.chen@ieee.org} \\ 
 }

\date{}

\begin{document}
\maketitle

\begin{abstract}
    The huge size of the widely used BERT family models has led to recent efforts about model distillation.
    The main goal of distillation is to create a task-agnostic pre-trained model that can be fine-tuned on downstream tasks without fine-tuning its full-sized version.
    Despite the progress of distillation, to what degree and for what reason a task-agnostic model can be created from distillation has not been well studied.
    Also, the mechanisms behind transfer learning of those BERT models are not well investigated either.
    Therefore, this work focuses on analyzing the acceptable deduction when distillation for guiding the future distillation procedure.
    Specifically, we first inspect the prunability of the Transformer heads in RoBERTa and ALBERT using their head importance estimation proposed by \citet{michel2019sixteen}, and then check the coherence of the important heads between the pre-trained task and downstream tasks.
    Hence, the acceptable deduction of performance on the pre-trained task when distilling a model can be derived from the results, and we further compare the behavior of the pruned model before and after fine-tuning.
    Our studies provide guidance for future directions about BERT family model distillation.\footnote{We will release all the scripts used once accepted.}
\end{abstract}

\section{Introduction}
Huge-sized BERT family pre-trained models \cite{devlin-etal-2019-bert,liu2019roberta,yang2019xlnet,conneau2019cross,lewis2019bart,raffel2019exploring} play an important role in many NLP applications. 
However, their large size has caused difficulty for both training and inference.
So there are many recent attempts to slim down a BERT family model by distillation~\cite{sun-etal-2019-patient,jiao2019tinybert,turc2019well,sanh2019distilbert,sun2020mobilebert,tsai-etal-2019-small}.

We care about the possibility of having a task-agnostic distilled model, which is more ideal for deep NLP practitioners.
A model is \emph{task-agnostic} if it can achieve competitive performance on downstream tasks by simply fine-tuning.
In contrast, most prior distilled models requires imitating from a fine-tuned full-size teacher model when transfer to downstream task, and thus they are not task-agnostic.
An exception is MobileBERT \cite{sun2020mobilebert}, which is a task-agnostic BERT distilled from a BERT-Large-sized BERT-like model with inverted-bottleneck.
However, MobileBERT still performs worse than its teacher model\footnote{MobileBERT is smaller and faster than BERT-Base.}.
Thus above efforts showed the difficulty of having a task-agnostic pre-trained model with performance competitive to its teacher model.
This difficulty motivates us to further investigate to what degree and for what reason a task-agnostic model is realizable.

In this work, we first check the prunablity of fine-tuned RoBERTa and ALBERT models.
The size of the downstream model can be viewed as a lower bound of a task-agnostic model. 
We use the attention head importance estimation to prune unimportant heads~\cite{michel2019sixteen}.
It is different from prior works that mainly focused on heads pruning of Transformer models trained from scratch~\cite{michel2019sixteen,voita-etal-2019-analyzing}.
Furthermore, 
our paper finds a better importance estimator for future usage (\S~\ref{sec:prunability}). 

We then provide insights for task agnostic distillation by comparing different behaviors between the pre-trained models and the fine-tuned models.
It is different from most previous studies on BERT \cite{rogers2020primer} that mainly focused only on the pre-trained models.
First, we investigate the consistency between the set of important heads in the pre-train task and that in the downstream tasks.
The consistency is vital, because the distillation loss encourages knowledge extraction from the heads important to pre-train task, and inconsistency may result in the poor performance during fine-tuning.
Our analysis in \S~\ref{sec:consistency} reveals the relation between \emph{head consistency} and the \emph{performance on the pre-train task}.
Second, we compare the outputs of the intermediate layers in the pre-trained and fine-tuned RoBERTa models.
Our results support that feature matching is a viable direction for BERT distillation (\S~\ref{sec:compare-behavior}).

\section{Experimental Settings}
In order to make our study more general, two different models on four different GLUE \cite{wang2018glue} tasks are investigated.
We mainly focus on RoBERTa-Large, because it is a BERT model pre-trained with the enhanced training objective and is thus representative.
Results of ALBERT-Base-v1 is also included due to its special parameters-sharing structure.
Four different tasks, grammaticality prediction, sentiment analysis, duplicated questions detection, and natural language inference, are picked for the experiments, where CoLA~\cite{warstadt2019neural}, SST-2~\cite{socher-etal-2013-recursive}, QQP\footnote{\url{https://www.quora.com/q/quoradata/First-Quora-Dataset-Release-Question-Pairs}} and MNLI~\cite{mnli2018} are utilized for the target tasks respectively.
For each model and each task, we train 5 models with different random seeds.

\section{Model Prunability}
\label{sec:prunability}
To examine the prunability of BERT models, we compute the importance of heads in the trained models with the assumption that not all heads are equally important~\cite{michel2019sixteen,voita-etal-2019-analyzing}.
We start with the notation introduction and the head importance estimation for better elaborate the conducted experiments. 

\subsection{Notation}

In this paper, we follow the notation in \citet{michel2019sixteen}.
For both RoBERTa and ALBERT, we denote the $N_h$-head attention function in the Transformer block at the $l$-th layer as
\begin{equation}
    \mathrm{MHAtt}^{(l)}(x) = 
    \sum_{h=1}^{N_h} \xi_h^{(l)} Att^{(l)}_{W_q^{(h,l)}, W_k^{(h,l)}, W_v^{(h,l)}}(x),
\end{equation}
where matrices $W_q^{(h,l)}, W_k^{(h,l)}, W_v^{(h,l)}$ are trainable parameters, and $\xi_h^{(l)}$ is the constant $1$.
For ALBERT, those parameters are shared across layer.
In the following descriptions, we abbreviate the $h$-th head in the $l$-th layer as the head $(h, l)$.
Because heads are shared across layers in ALBERT, the layer it belongs to does not need to be specified.

\subsection{Head Importance Estimation}
In order to shrink the model, we prune the heads that have lower importance, where we compute the head importance estimation proposed by \citet{michel2019sixteen}.
For RoBERTa, the importance of the head $(h, l)$ is calculated as
\begin{equation}
    I_h^{(l)} = \mathbb{E}_{x \sim X} 
    \abs{ \partial L(x) / \partial \xi^{(l)}_h}.
    \label{eq:head-score-robeta}
\end{equation}
For ALBERT with $L$ layers, since the parameters of heads are shared across layers, the importance of head $h$ is calculated as 
\begin{equation}
    I_h = \mathbb{E}_{x \sim X} 
    \abs{\sum_{l=1}^L \partial L(x) / \partial \xi^{(l)}_h}.
    \label{eq:head-score-albert}
\end{equation}
In practice, the expectation in (\ref{eq:head-score-robeta}) and (\ref{eq:head-score-albert}) are estimated by averaging over samples in the training dataset. 
For RoBERTa, the importance scores are normalized for each layer:
\begin{equation}
    \bar{I}_h^{(l)} = I_h^{(l)}/ (\mathrm{Norm}( I_1^{(l)}, \cdots, I_H^{(l)} )),
    \label{eq:head-score-norm}
\end{equation}
where $H$ is the number of heads
Different normalization methods are performed, and we discuss their difference in the next section.

\subsection{Prunability of RoBERTa and ALBERT}

\begin{figure*}[t!]
    \centering
    \includegraphics[width=0.24\linewidth]{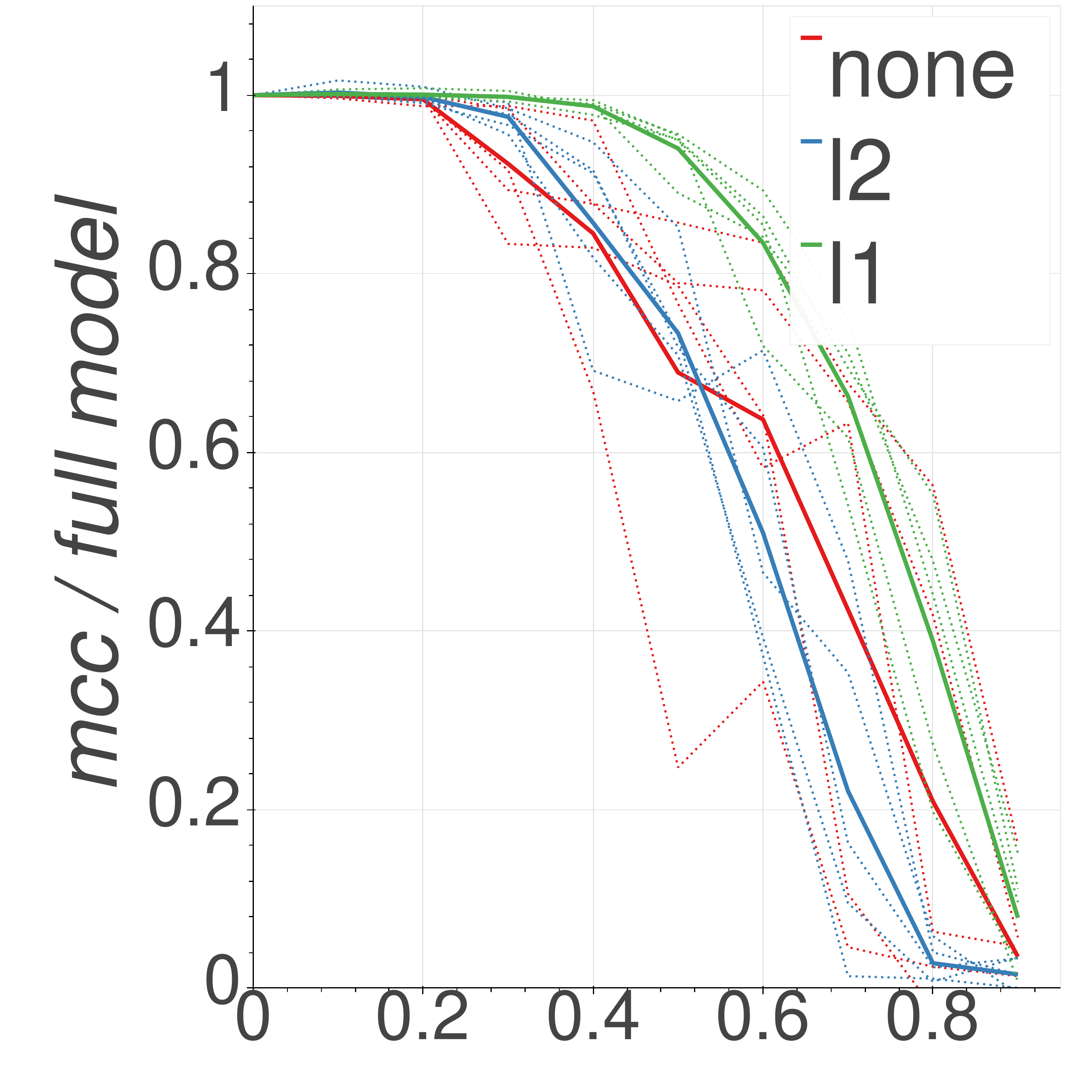}
    \includegraphics[width=0.24\linewidth]{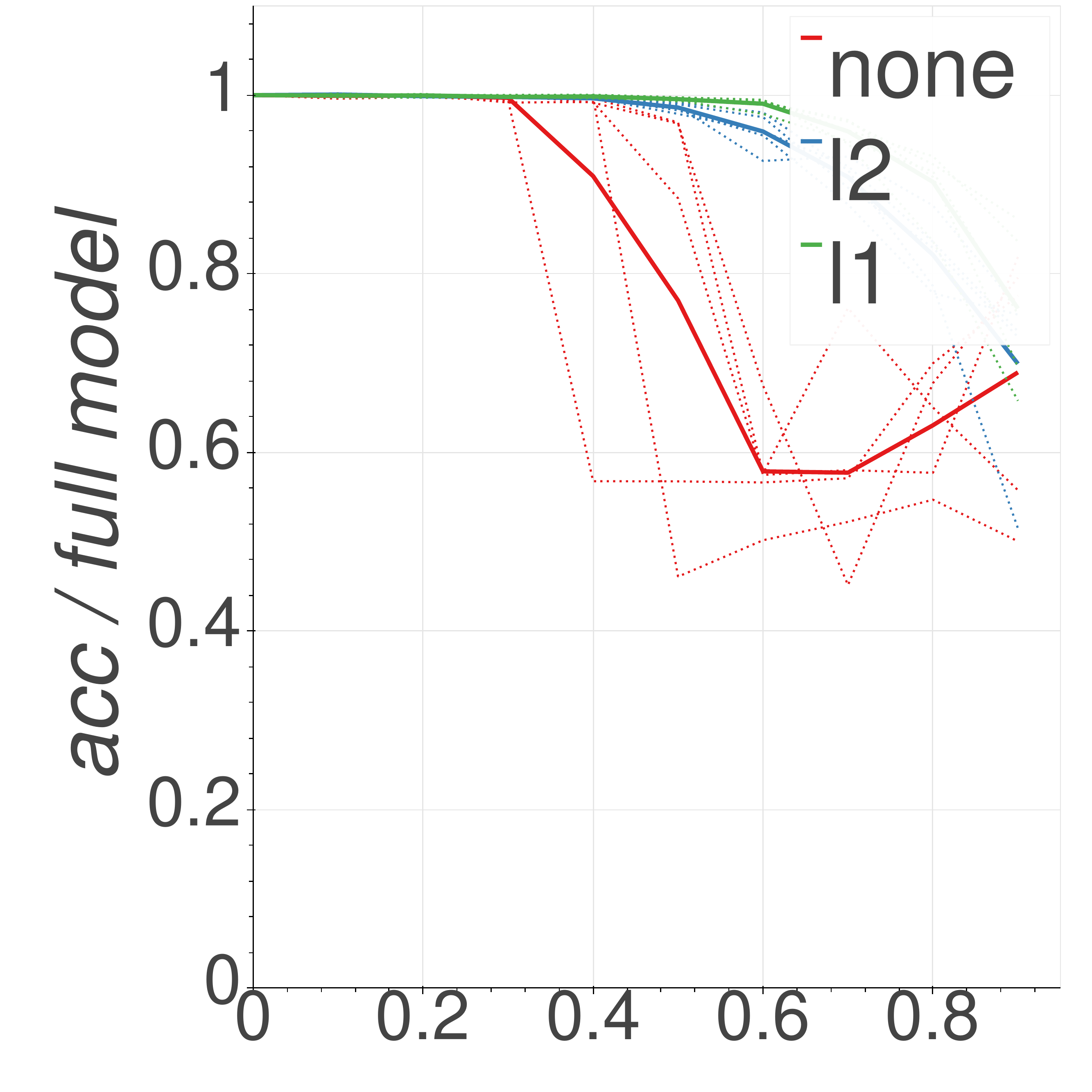}
    \includegraphics[width=0.24\linewidth]{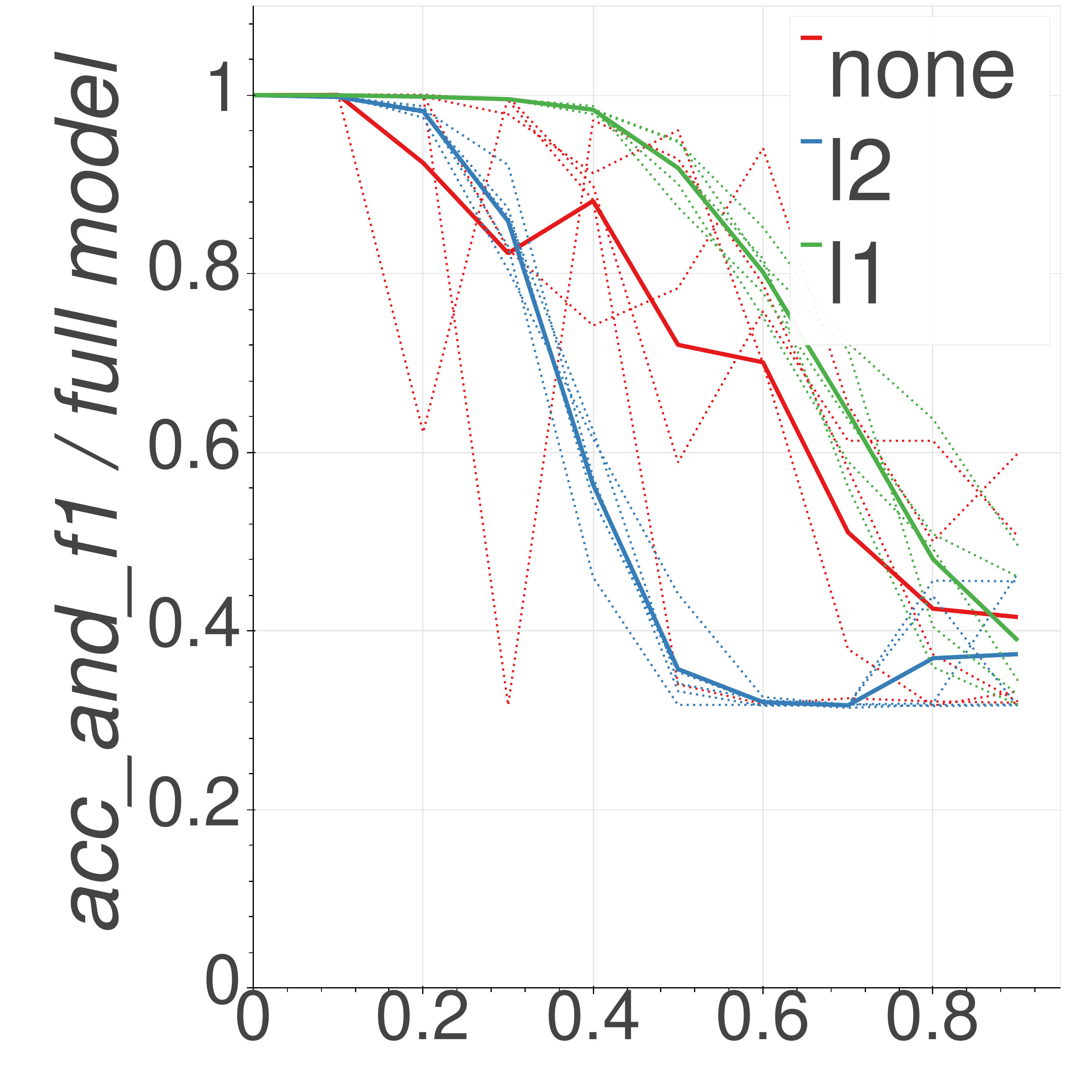}
    \includegraphics[width=0.24\linewidth]{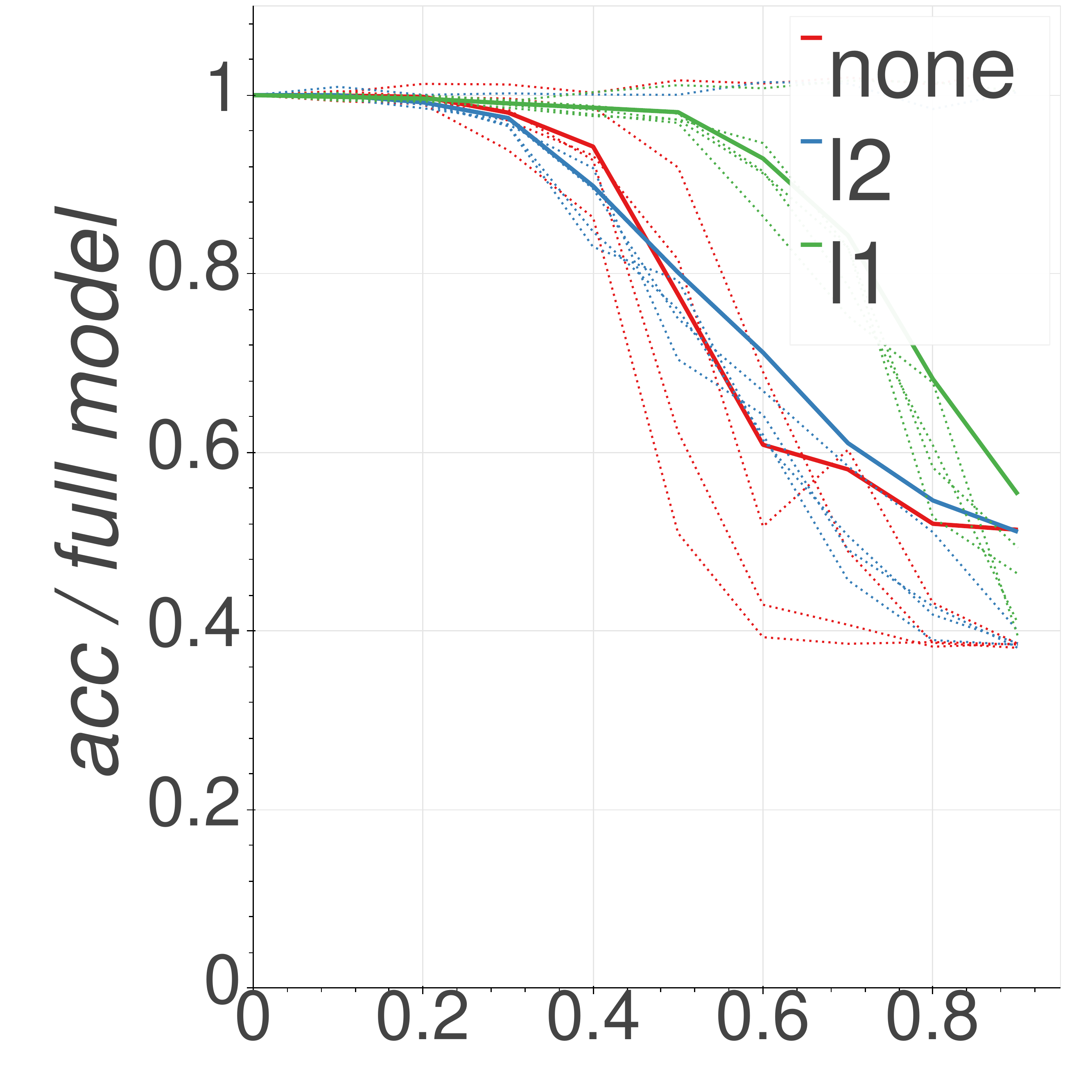}
    \includegraphics[width=0.24\linewidth]{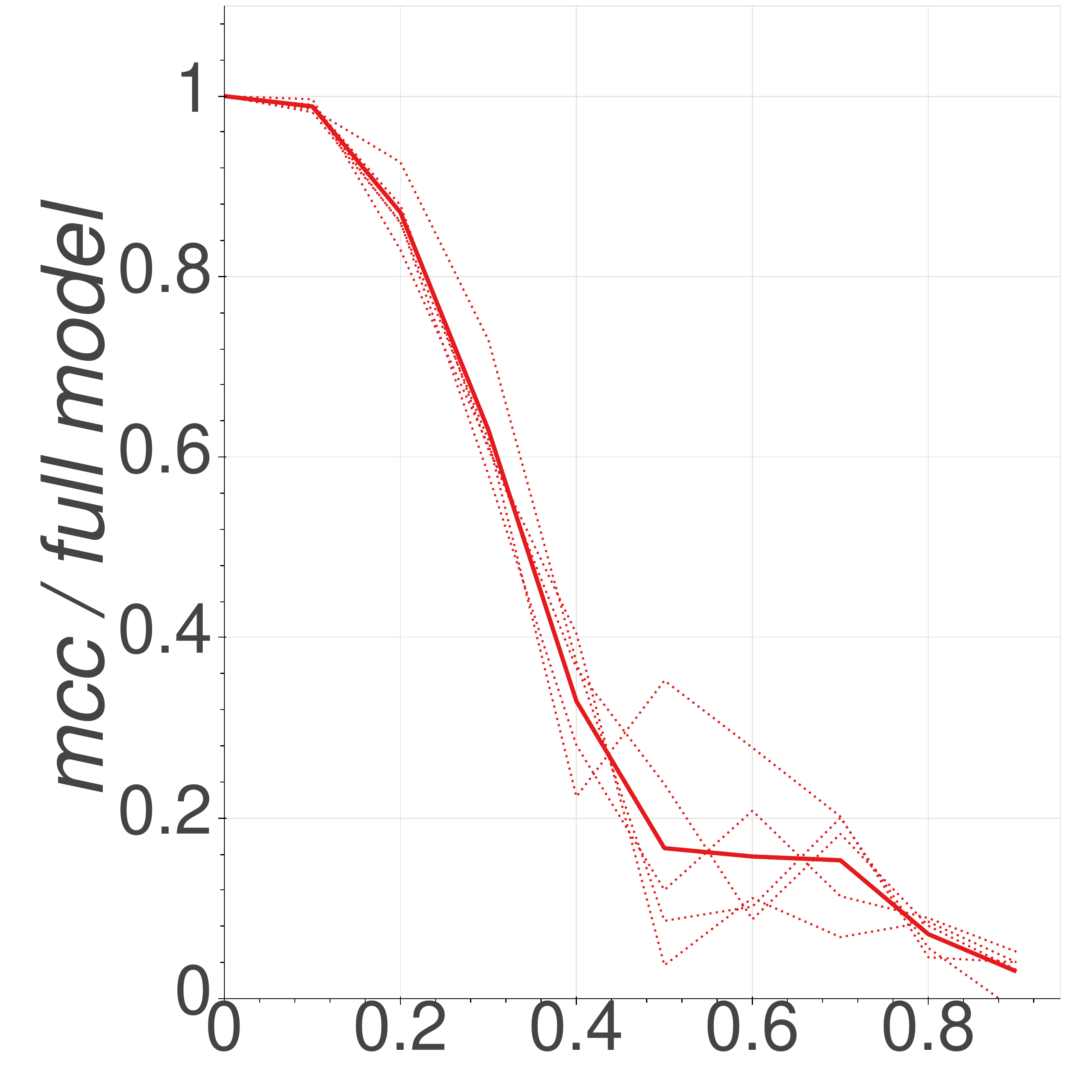}
    \includegraphics[width=0.24\linewidth]{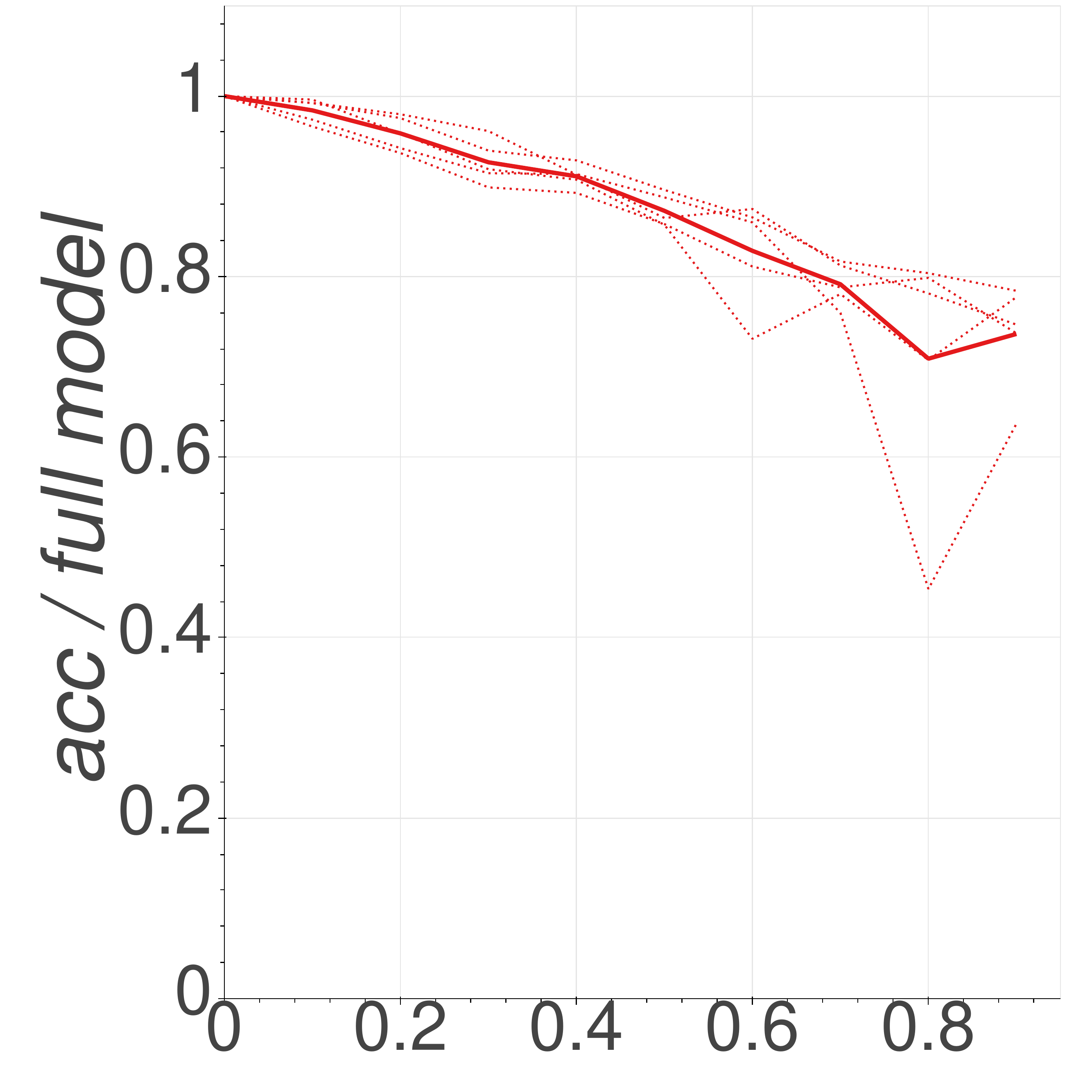}
    \includegraphics[width=0.24\linewidth]{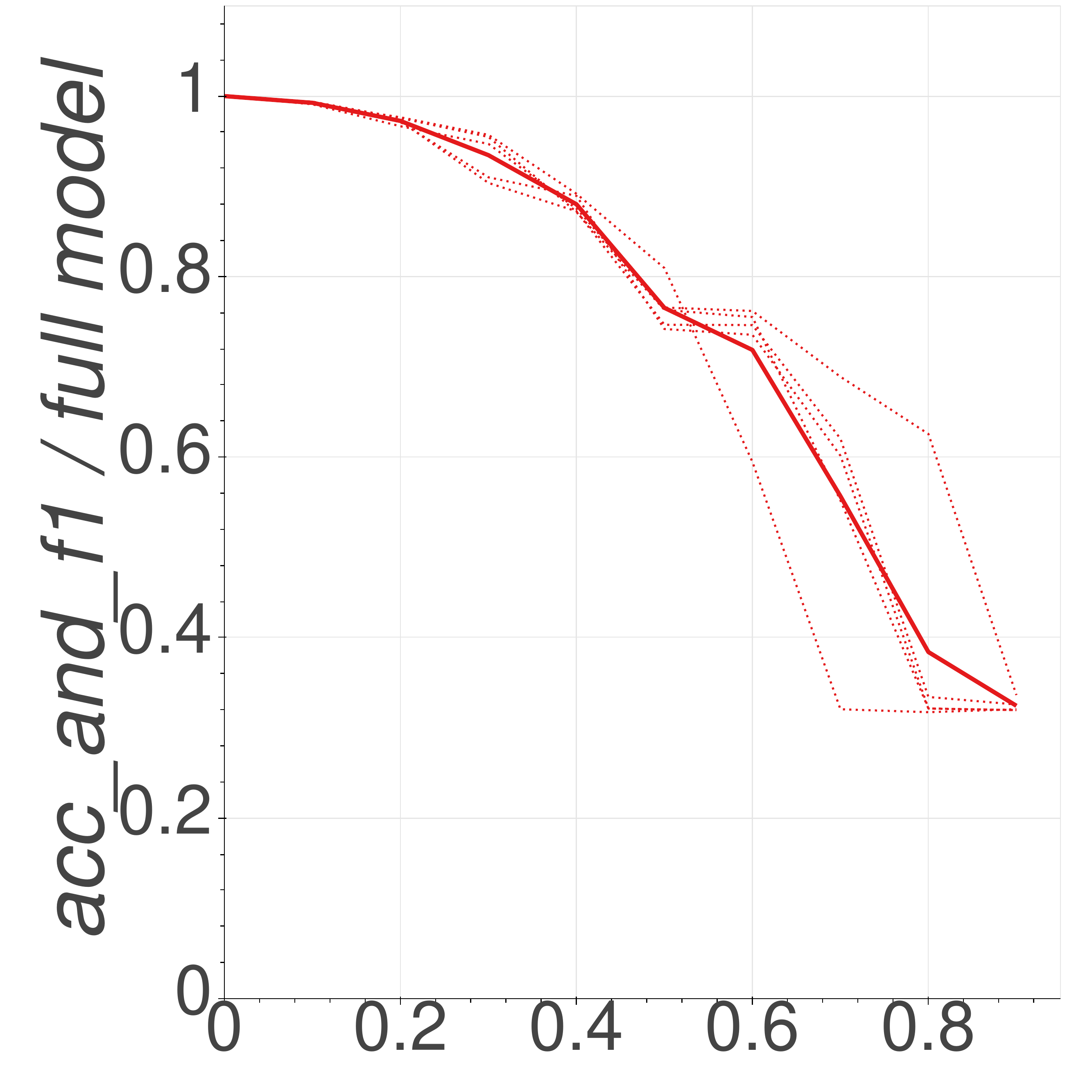}
    \includegraphics[width=0.24\linewidth]{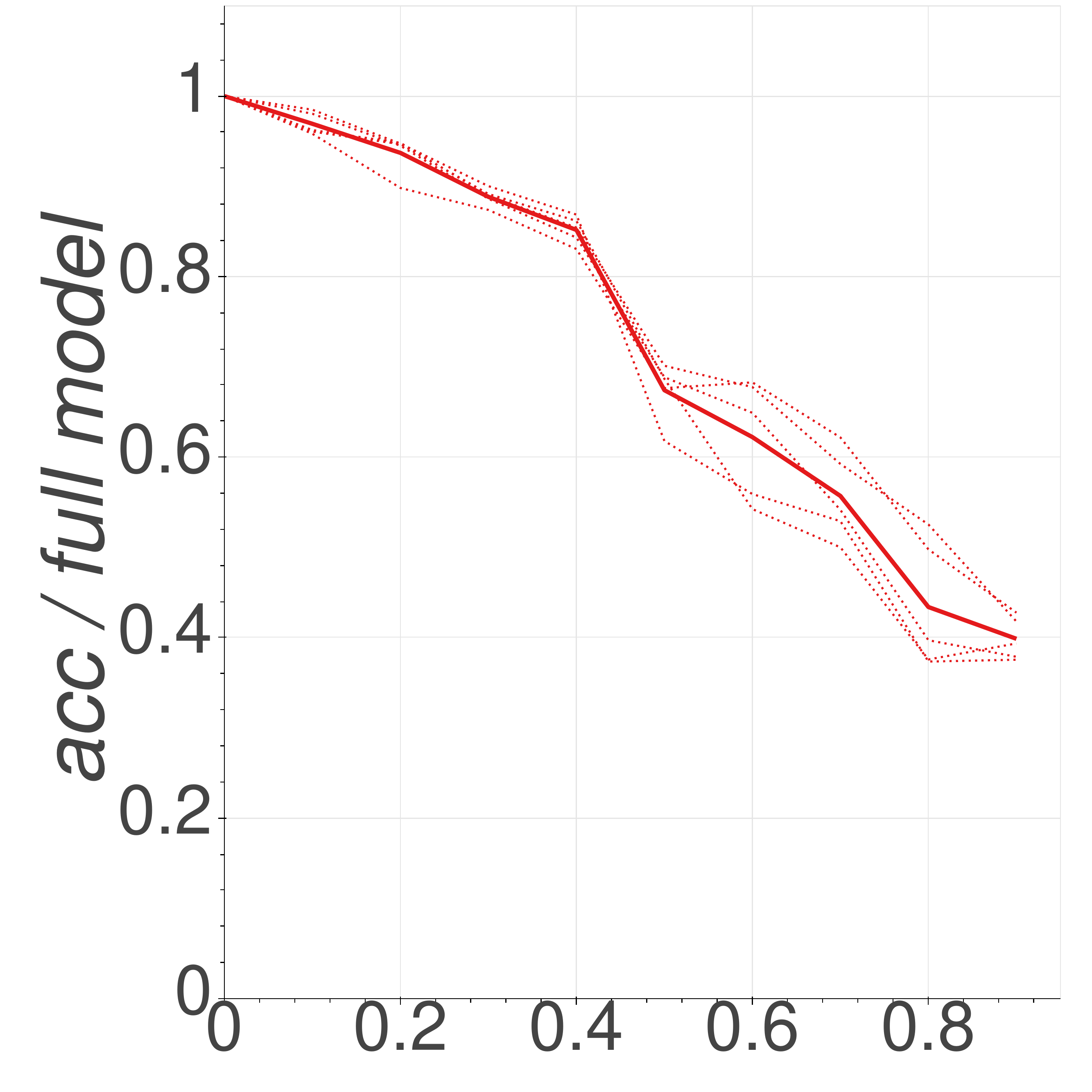}
    \caption{Relative performance of pruned models on downstream tasks. The x-axis indicates the ratio of heads pruned. The y-axis indicates the performance relative to the full model. Dotted lines represent results with different random seeds.
    The two rows are the results of RoBERTa and ALBERT, and the four columns are the performance of CoLA, SST-2, QQP, MNLI, respectively.}
    \label{fig:pruning-acc}
    \vspace{-3mm}
\end{figure*}


In this paper, the normalization function used in (\ref{eq:head-score-norm}) is further investigated.
\citet{MolchanovTKAK17} showed that normalizing with $l_2$ norm empirically perform well.
Inheriting from \citet{MolchanovTKAK17}, $l_2$ norm is also adopted in \citet{michel2019sixteen}.
Note that we argue the discrepancy between our settings and the previous ones, so performing all normalizations in our setting can make the experiments complete and the derived conclusions convincing.
Therefore, using $l_1$ norm, $l_2$ norm, and no normalizing are conducted in the experiments.

\begin{figure}[t!]
    \centering
    \includegraphics[width=0.475\linewidth]{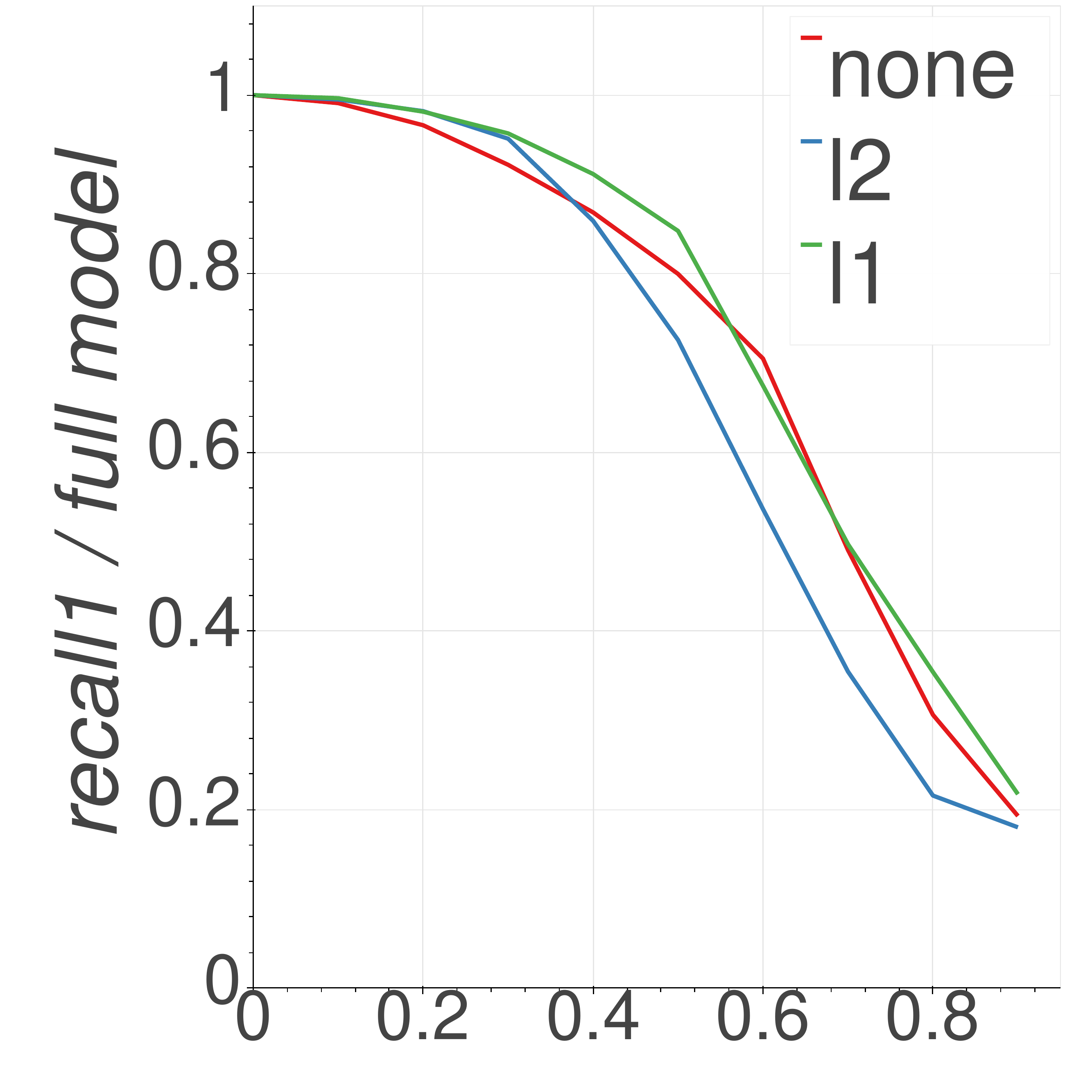}
    \includegraphics[width=0.475\linewidth]{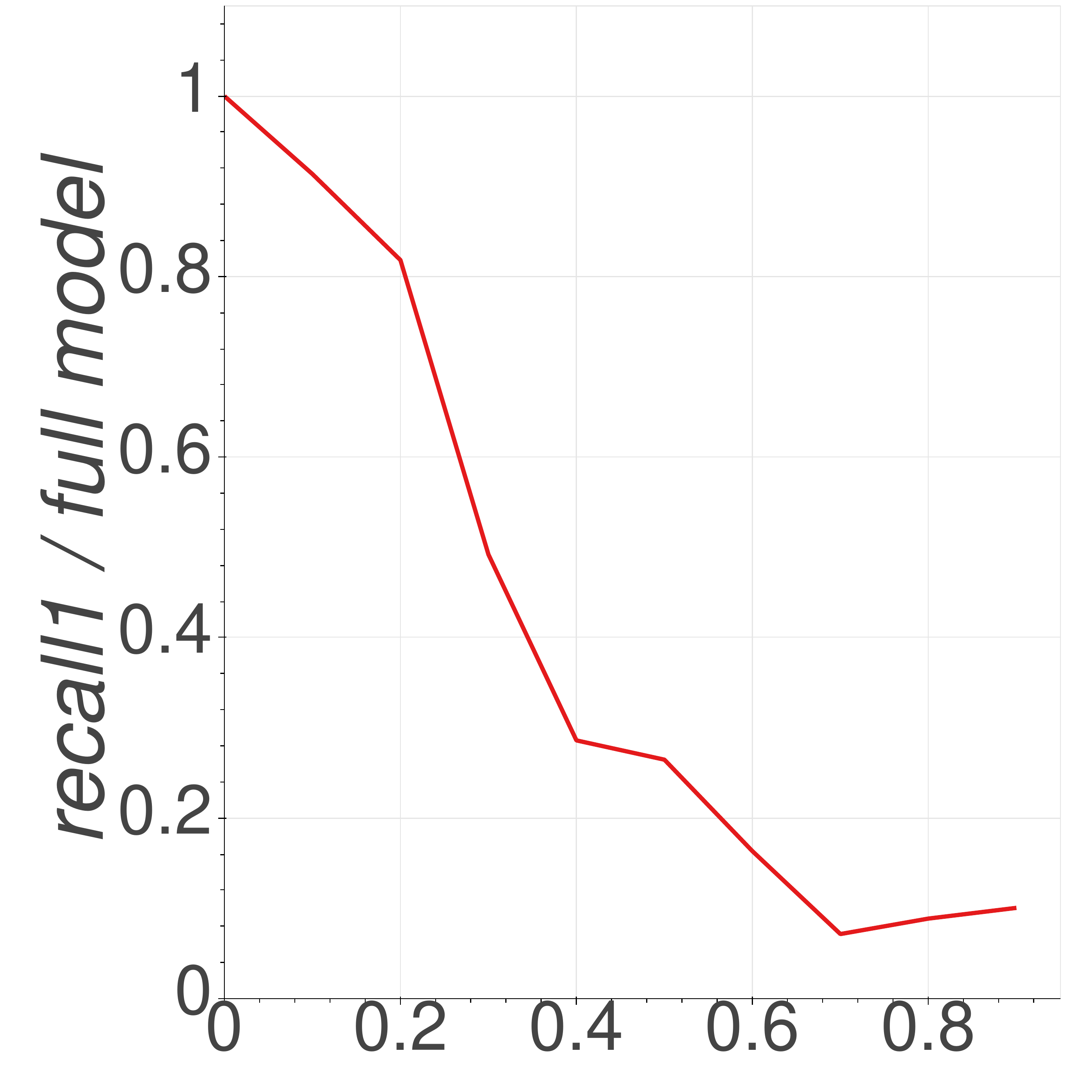}
    \caption{Performance of RoBERTa (left) and ALBERT (right) on pre-train task (masked LM). The x-axis is the ratio of the heada pruned.}
    \label{fig:pruning-wiki}
    \vspace{-3mm}
\end{figure}

Following \citet{michel2019sixteen}, we prune the attention heads iteratively.
According to the estimated importance, we prune the heads starting from the least important ones to the most important ones and evaluate the performance for different ratios of pruned heads.
Figure~\ref{fig:pruning-acc} and \ref{fig:pruning-wiki} show the relation between the performance relative to the unpruned models and the pruned ratios for downstream and pre-trained tasks respectively.

Figure~\ref{fig:pruning-acc} shows that RoBERTa may contain more redundancy to slim down than ALBERT.
RoBERTa consistently retains over 90\% performance when pruning more than 50\% heads for all tasks.
In comparison, performance of ALBERT drops more quickly, especially on CoLA and MNLI.
It may be reasonable, since ALBERT here has only 12 heads while the RoBERTa-Large has 384 heads.\footnote{The same phenomenon is observed for ALBERT-XXLarge-v2 with 64 heads. Results are included in Appendix.}
We thus focus on RoBERTa for the following analysis.

Figure~\ref{fig:pruning-acc} also shows that normalizing with $l1$ norm is more effective.
Pruning with $l1$ norm can retain higher performance when the same ratio of heads are pruned.
It also implies that normalizing with $l1$ better reflects the head importance.
Therefore, we use $l1$ norm for estimating head importance in the following study.

\section{Important Head Consistency between Pre-Train and Downstream Tasks}
\label{sec:consistency}

With the head importance estimation as a tool, we check the consistency of the important heads between pre-train and  downstream tasks.
As a multi-head attention model is distilled, the knowledge from more important heads should be more likely to be extracted, while the knowledge in the less important heads may lose.
The distilled model may fail to transfer the knowledge to a downstream task if less important heads in the pre-train task are important to the downstream task.
Hence, higher consistency can ensure the potential of performing well in the downstream task.

Such consistency can be quantified with the recall metrics of the heads.
It quantifies what ratio of the heads important for the downstream task is retained when the model is pruned on pre-trained task. 
Specifically, we define \textit{recall} as:
\begin{equation}
    \mathrm{Recall}_{0.9}(x) = \abs{H_x^{(p)} \cap H_{0.9}^{(d)}} / \abs{H_{0.9}^{(d)}},
    \label{eq:recall}
\end{equation}
where $H^{(p)}_x$ is the head set required to achieve $x\%$ relative performance on the pre-train task,
and $H^{(d)}_{0.9}$ is the head set required to achieve $90\%$ relative performance on a downstream task. 

\begin{figure}
    \centering
    \includegraphics[width=0.80\linewidth]{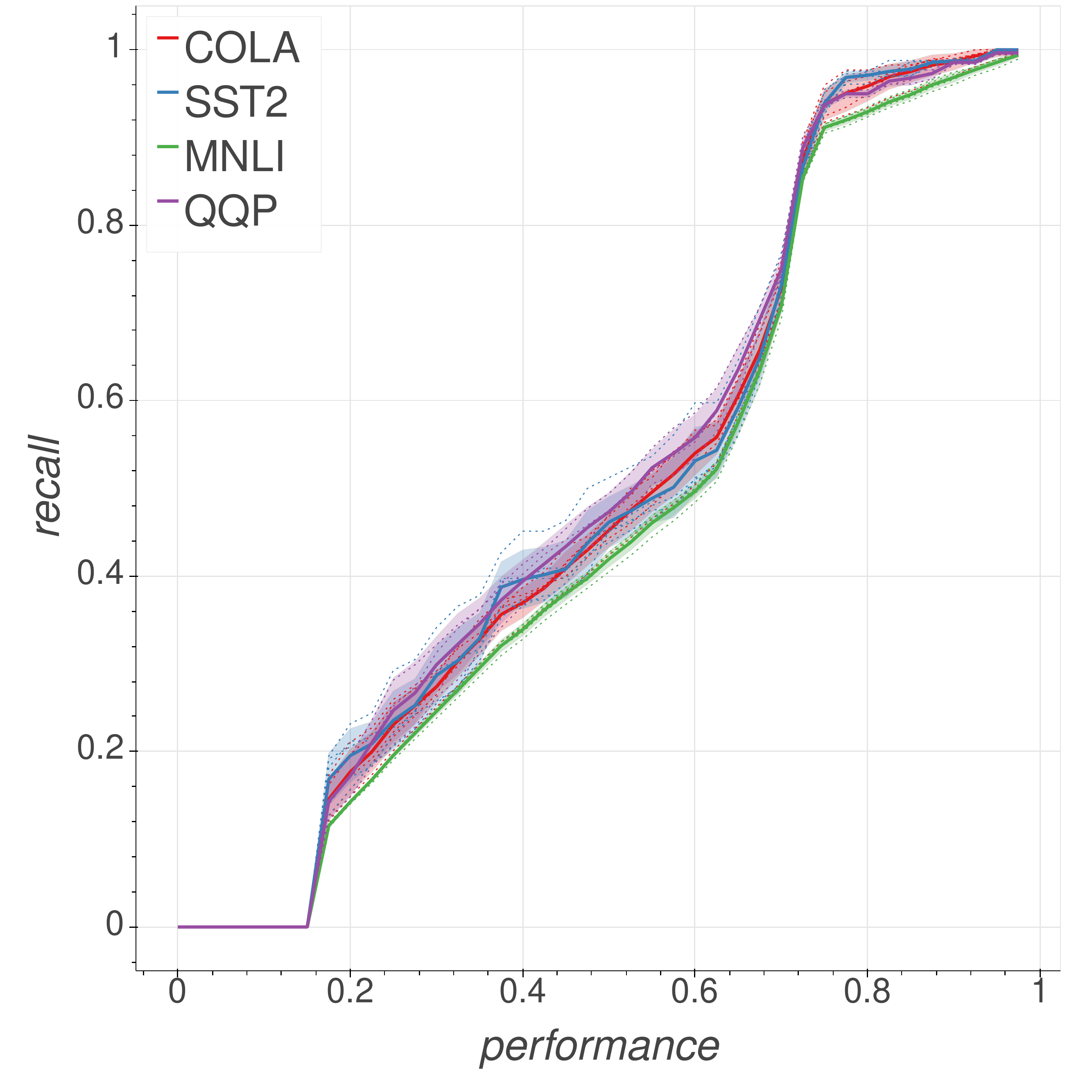}
    \caption{Recall defined in (\ref{eq:recall}). The shaded area represents the standard deviation.}
    \label{fig:roberta-common}
\end{figure}

Figure~\ref{fig:roberta-common} plots the \textit{recall} curve and shows that the high recall (over 90\%) can be achieved when the relative performance on the pre-train model is around 75\% for all 4 tasks.
Previous studies \cite{frankle2018the,prasanna2020bert} showed that an important sub-network with good initialization is sufficient to perform well on downstream tasks.
Combined with those studies, our results provide a guideline for model distillation: \textbf{the performance on downstream task is better guarantee when the distilled model retains 75\% relative performance on the pre-train task}.
Also, according to Figure~\ref{fig:pruning-wiki}, less than 50\% heads are required to maintain 75\% performance, implying that RoBERTa may be slimmed down by at least 50\%.

\section{Behavior Comparison of Pre-Trained and Fine-Tuned Models}
\label{sec:compare-behavior}

To better understand how RoBERTa learns, we directly compare the attention weights and the intermediate outputs in the pre-trained model and the model fine-tuned on the downstream task.
We first feed the same input from the downstream task to both pre-trained and fine-tuned models.
Then for each pair heads from the same location in these models, we measure the JS divergence \cite{lin1991divergence} between their attention weights on tokens.
Compared to \cite{peters2019tune}, it more directly reflects how static each head is throughout the process of fine-tuning.

    

\begin{figure}
    \centering
    \includegraphics[width=1\linewidth]{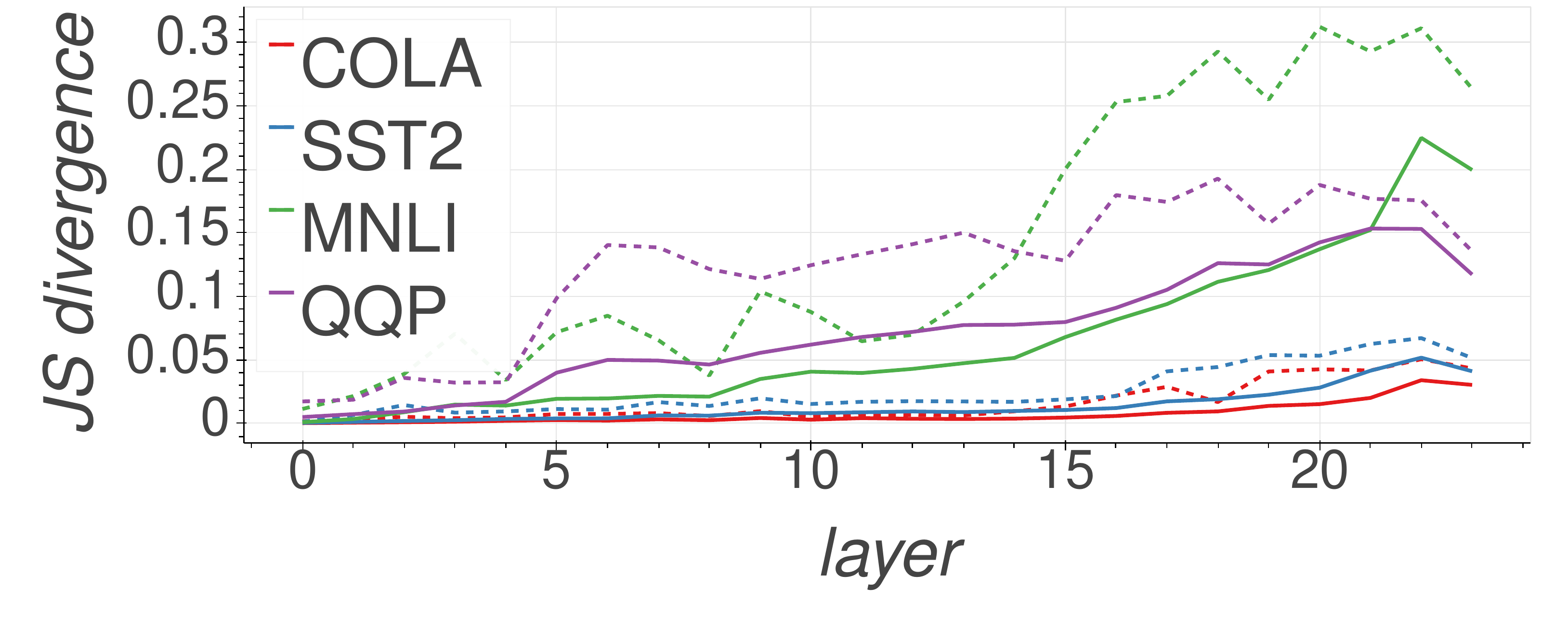}
    \includegraphics[width=1\linewidth]{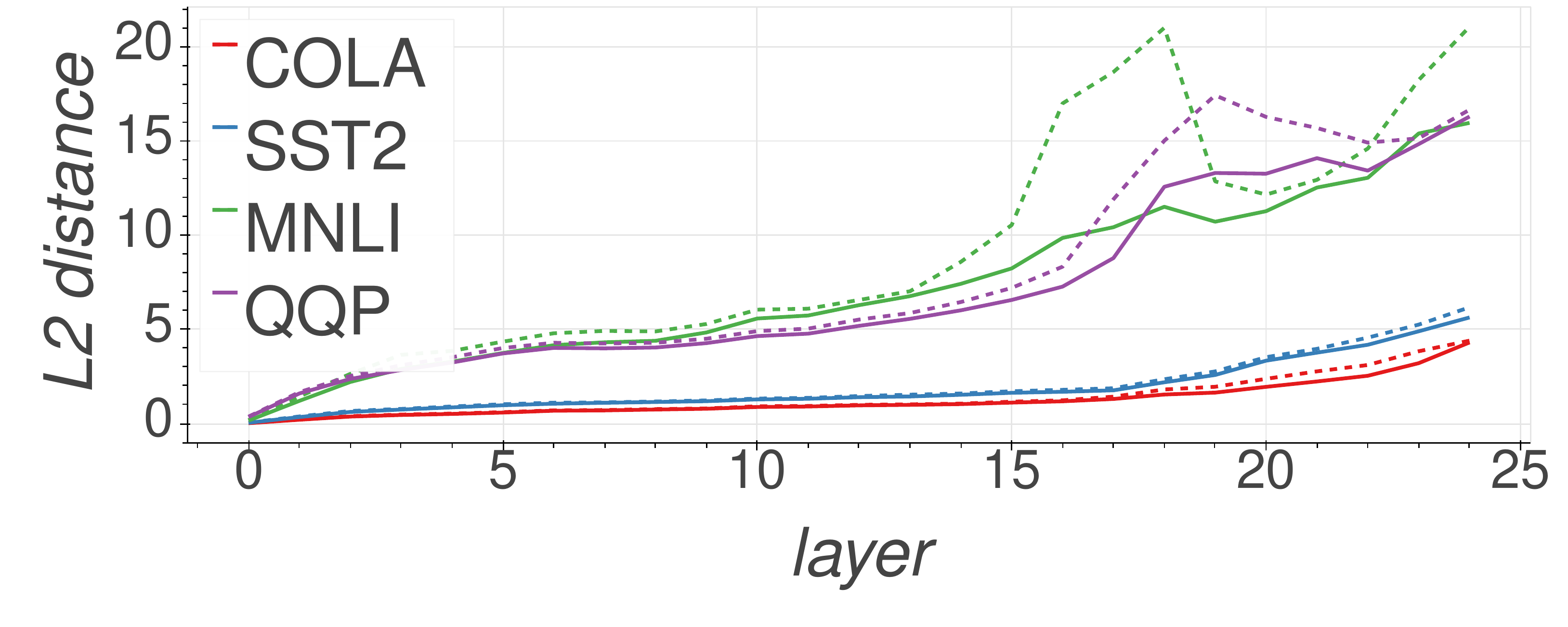}
    \caption{JS divergence between head weights of pre-trained and fine-tuned models (top). L2 distance between intermediate features from pre-trained and fine-tuned models (bottom). Average values are first calculated for each head/seed. The solid/dashed line are the average/max value for each layer.}
    \label{fig:compare-feature}
\end{figure}

The results are shown in Figure~\ref{fig:compare-feature}.
The change in the lower layers are small in general, implying that \textbf{lower-level features required for the pre-train task is coherent with the features for downstream tasks}.
To further validate the conclusion, we further experiment freezing the lower layers while fine-tuning RoBERTa.
As a preliminary experiment, we freeze 5 layers for all tasks.
As a result, the partly frozen model achieves relative performance 0.91 on CoLA, and more than 0.98 on the reset three tasks.
It shows that almost the same performance can be achieve despite the frozen layers.
Therefore, the common distillation practice that mimicking the output and attention of the lower layers in RoBERTa is indeed reasonable.

We also measure the correlation between the JS-divergence discussed in \ref{sec:compare-behavior} and the head importance head importance estimated by \ref{eq:head-score-norm}.
Surprisingly, for both the pre-train task and the downstream task, the head importance have little correlation ($\leq$ 0.10 for all tasks) with the JS-Divergence.
The roles those heads play during the process of fine-tuning remain to further investigate.

\section{Conclusion}
In this paper, we first inspect the prunability of both RoBERTa and ALBERT and discover that normalizing with $l1$ norm makes the importance estimation better applicable for RoBERTa.
Also, our studies provide the guideline about the acceptable performance deduction on the pre-train task during the distillation.
Based on the empirical results, this paper reveals that RoBERTa is good at not only initializing parameters but also extracting features, aligning with the common distillation practice.
Our complete studies provide benefits for future research on BERT family model distillation.

\bibliographystyle{acl_natbib}
\bibliography{anthology,acl2021}

\appendix
\section{Implementation Detail}
 Implementation by Hugging Face \footnote{\url{https://huggingface.co/}} is used for the downstream GLUE tasks and the masked language model pre-train task. Link to our code will be available once the paper is accepted.

\section{Computing Infrastructure}
Most of the experiments are executed with NVIDIA RTX 2080Ti, while some are with GTX 1080, and some are with NVIDIA Tesla P40. Despite a P40 is used, all of the experiments can be reproduced with a RTX 2080Ti GPU on a single entry-level workstation (even personal computer is possible).

\section{Dataset and Metrics}

The size of GLUE datasets and the metrics used are listed in table~\ref{tab:glue-info}.
They are same as the original settings, and can be download from the GLUE website \footnote{\url{https://gluebenchmark.com/tasks}}.
We use the implementation of the metrics by Hugging Face \footnote{\url{https://github.com/huggingface/transformers/blob/v2.8.0/examples/run_glue.py}}.
For the task MNLI and QQP, not all training data is used in each run for the sake of execution time.
20\% randomly sampled subset from the training data is used for the head importance estimation; 10\% randomly sampled subset are used for the prunability experiments in section~\ref{sec:prunability}; 25\% randomly sampled subset are used for the comparison in section~\ref{sec:compare-behavior}.

For the pre-train task, we used WikiDump extracted with the script \footnote{\url{https://github.com/saffsd/wikidump}}.
4649706 articles are extracted.
Among them, we randomly sample 10\% for head importance estimation, and the prunability experiments in section~\ref{sec:prunability} for the sake of execution time.
Recall@1 is used as the metric on the masked language modeling task.

\begin{table}[t!]
    \centering
    \begin{tabular}{lrrl}
        \toprule
        Task & \#Training & \#Dev & Metric \\
         \midrule
         CoLA & 8,551 & 1,042 & MCC \shortcite{matthews1975comparison} \\ \hline
         SST-2 & 67,350 & 873 & Acc. \\ \hline
         QQP & 363,847 & 40,431 & Avg Acc \& F1 \\ \hline
         MNLI & 392,703 & 9,,816 & Acc. \\
         \bottomrule
    \end{tabular}
    \caption{The data statistics and evaluation metrics.}
    \label{tab:glue-info}
\end{table}

\section{Results of ALBERT-XXLarge}
\begin{figure}[t!]
    \centering
    \includegraphics[width=0.49\linewidth]{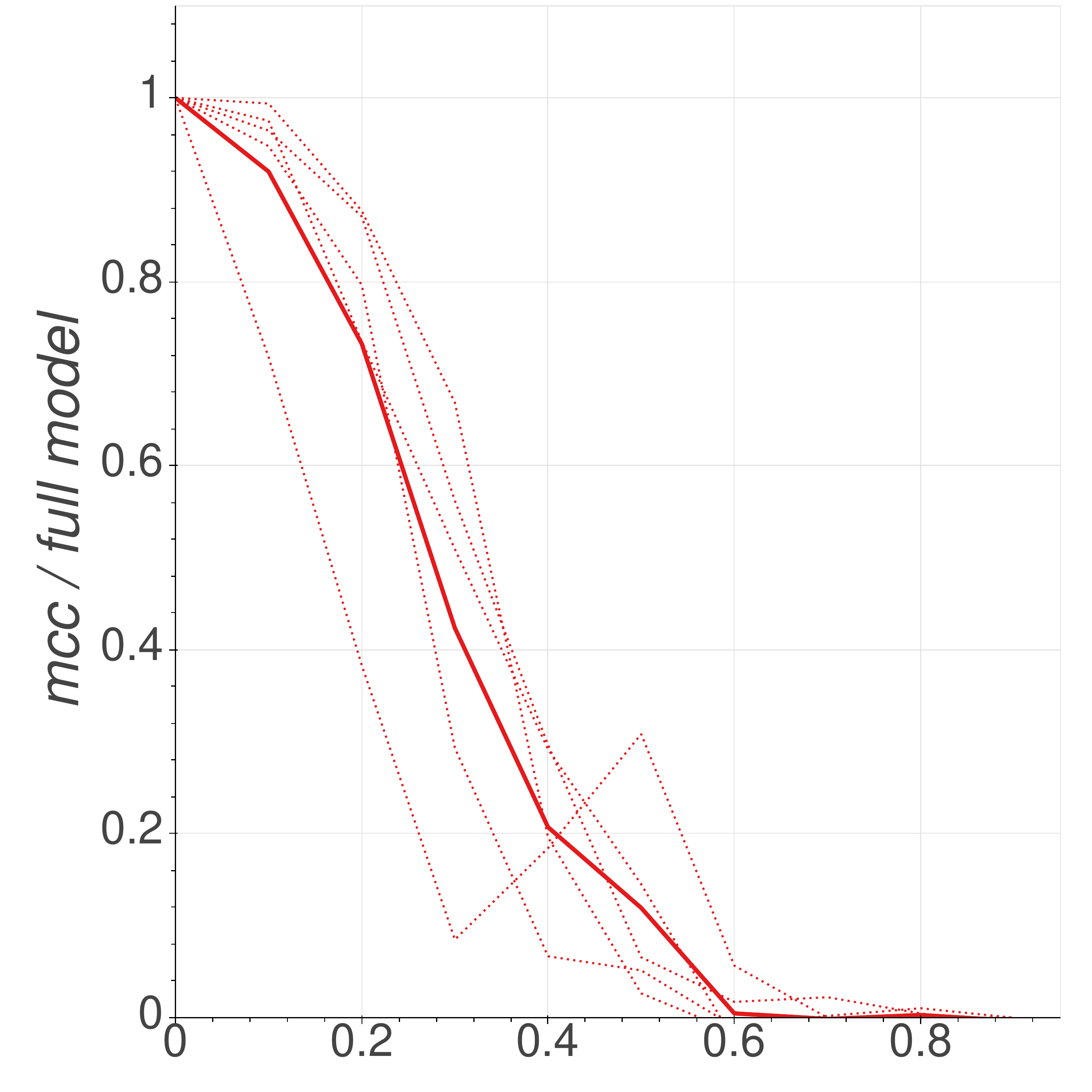}
    \includegraphics[width=0.49\linewidth]{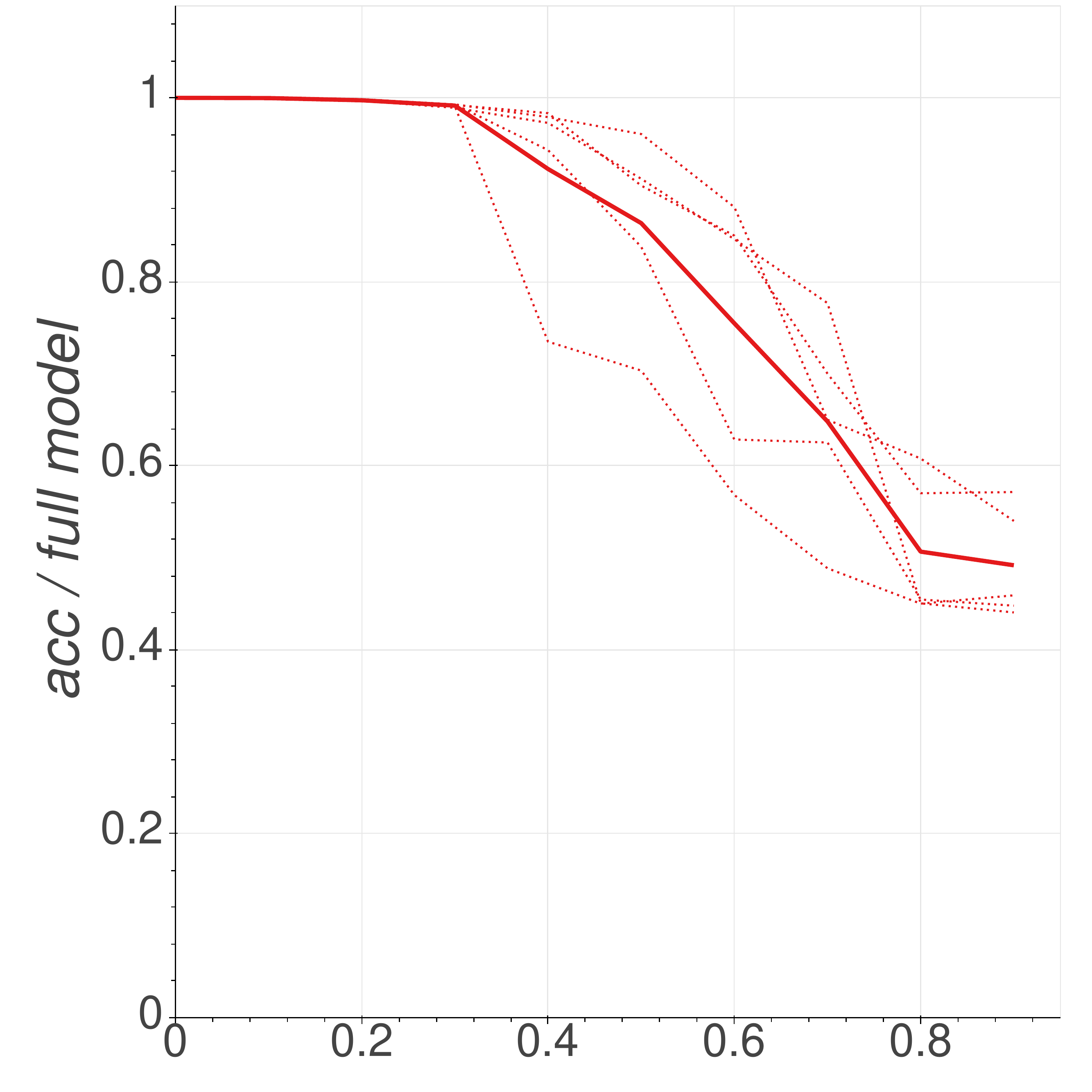}
    \includegraphics[width=0.49\linewidth]{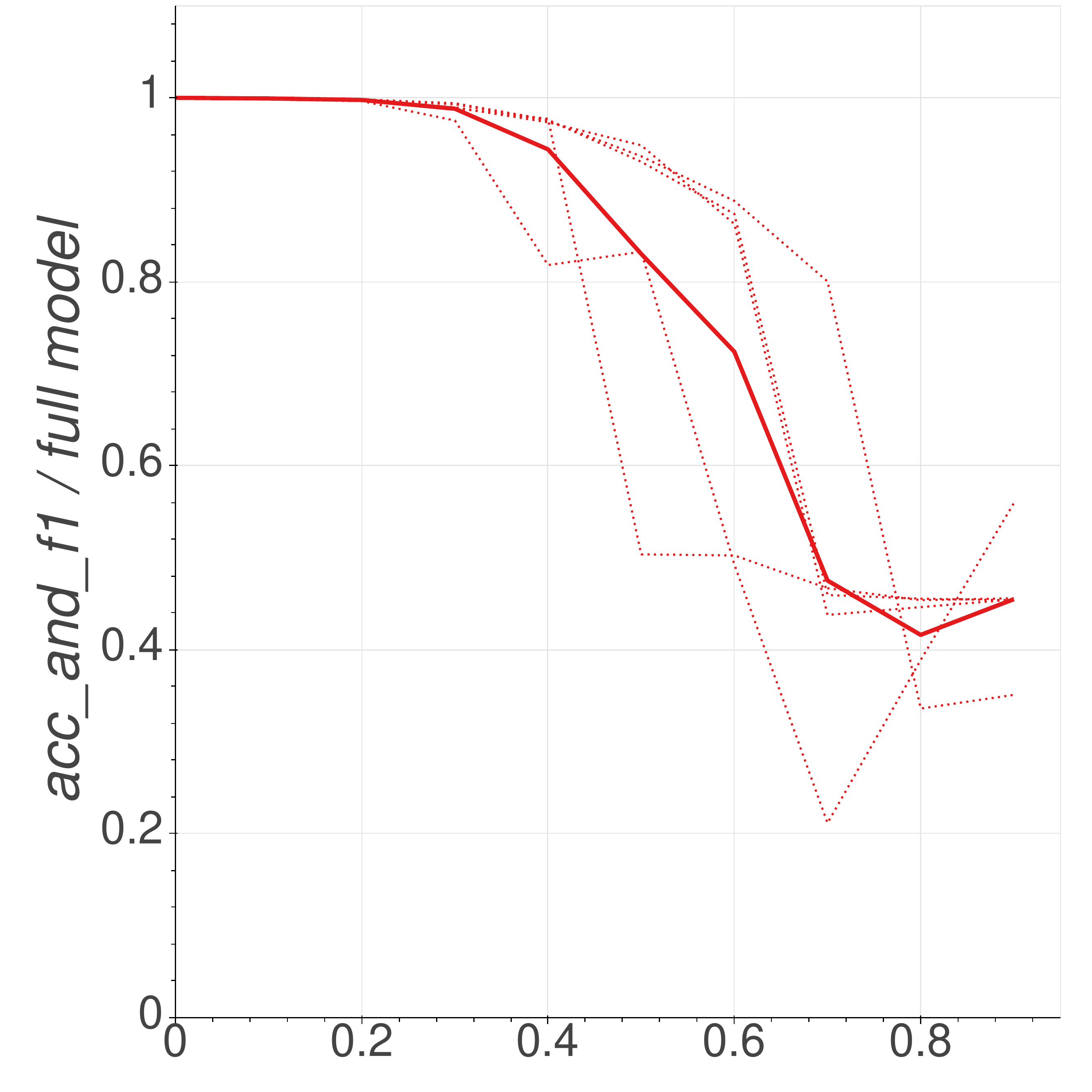}
    \includegraphics[width=0.49\linewidth]{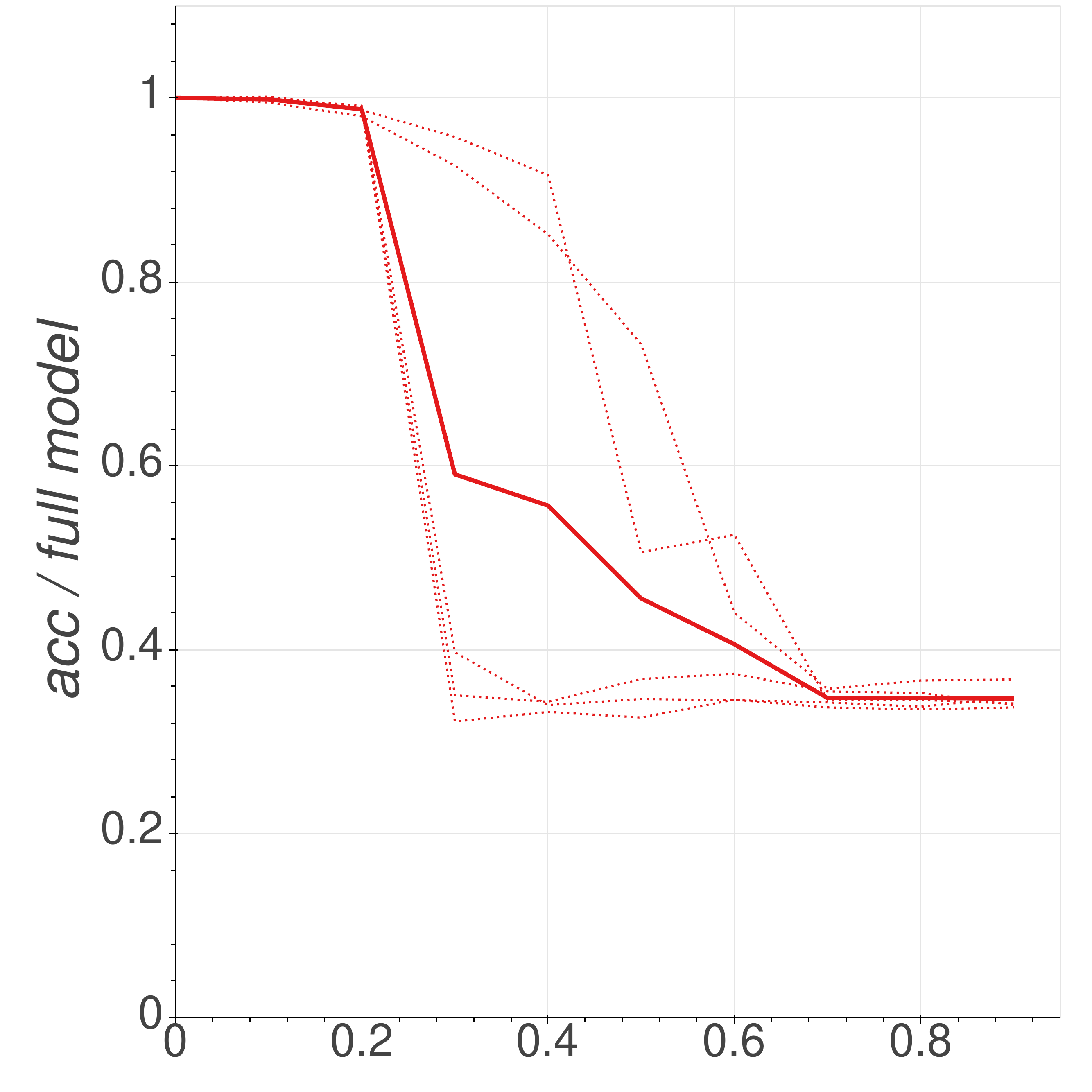}
    \caption{Relative performance of pruned ALBERT-XXLarge-v2 model on downstream tasks. The x-axis indicates the ratio of heads retained. The y-axis indicates the performance relative to full model. Dotted lines represent performance of model trained with different random seeds. The figures from left to right, top to down, are performance of task CoLA, SST-2, QQP, MNLI, respectively.}
    \label{fig:pruning-acc-albert-xxlarge}
\end{figure}

The results of the prunability experiments on ALBERT-XXLarge is in \ref{fig:pruning-acc-albert-xxlarge}.
It shows that ALBERT-XXLarge is also less prunable than RoBERTa-Large.

\section{Correlation between Head Importance and JS-Divergence}
\begin{figure*}
    \centering
    \includegraphics[width=0.24\linewidth]{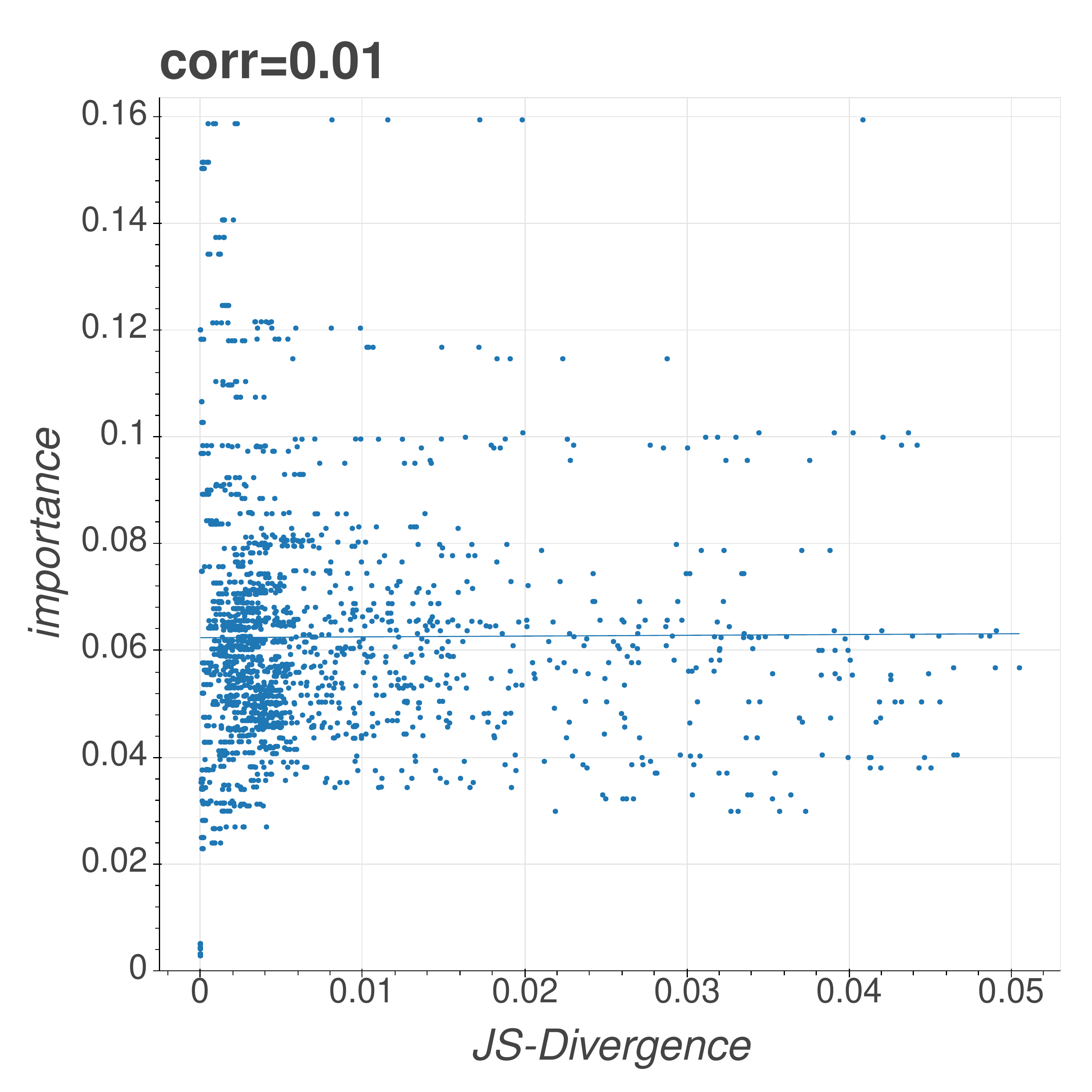}
    \includegraphics[width=0.24\linewidth]{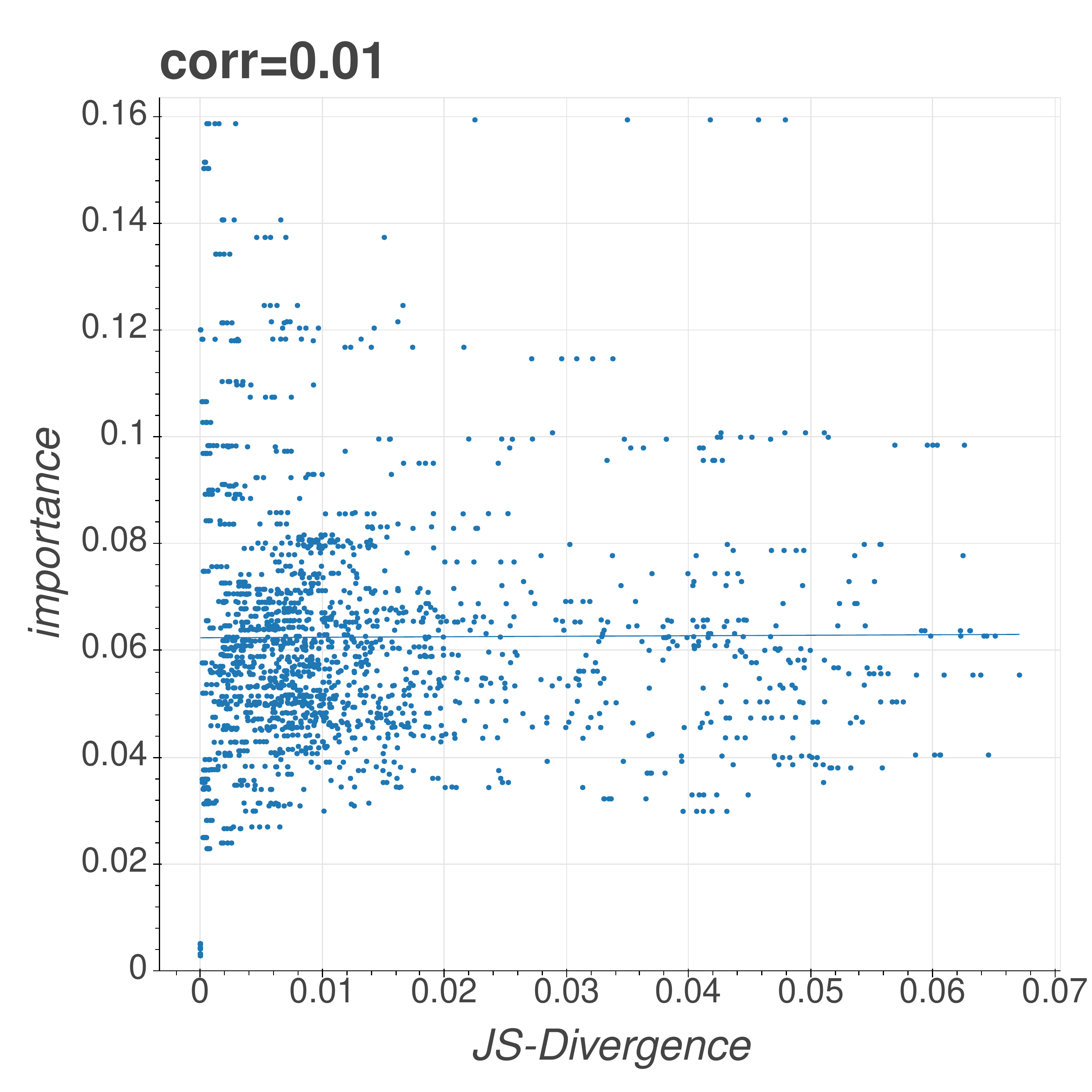}
    \includegraphics[width=0.24\linewidth]{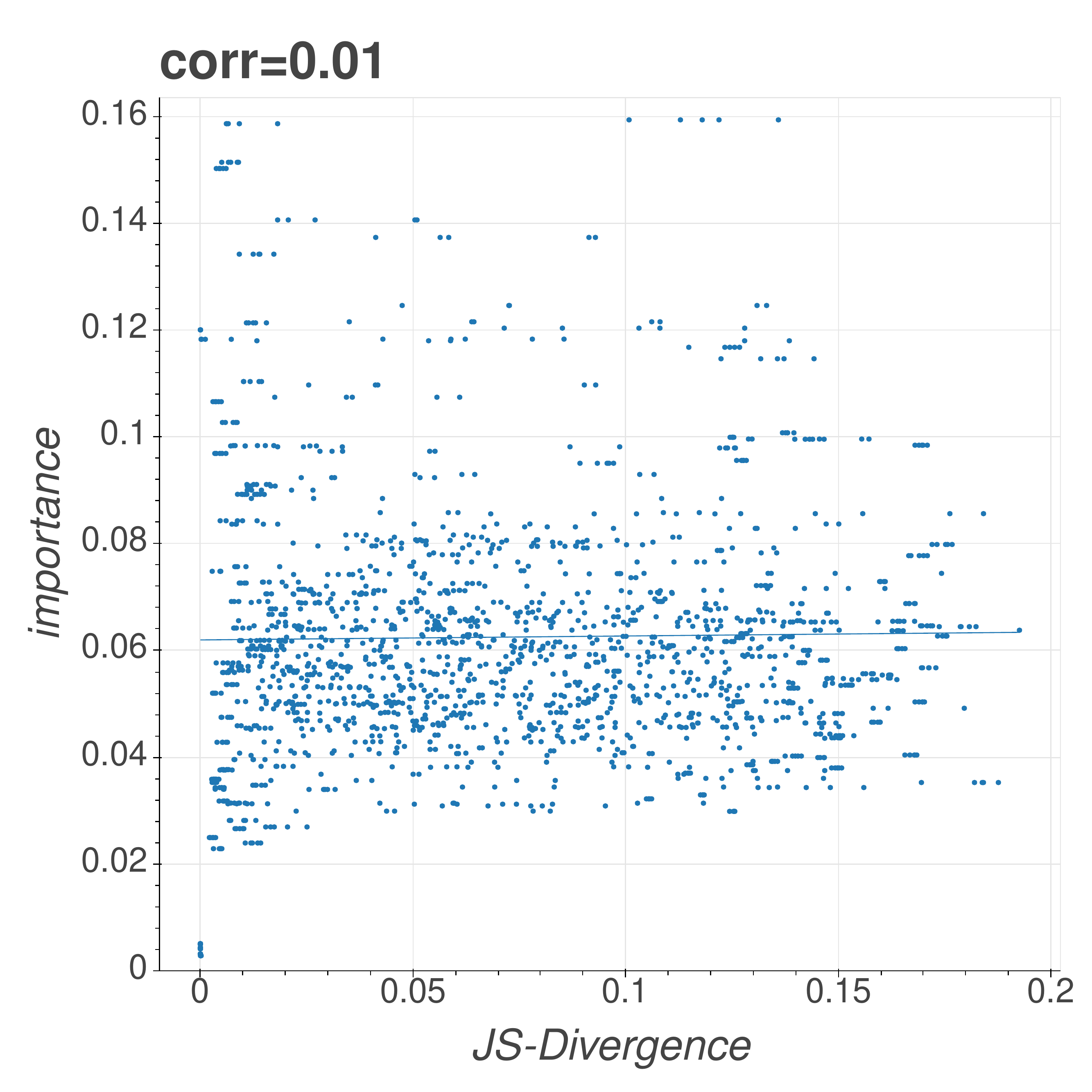}
    \includegraphics[width=0.24\linewidth]{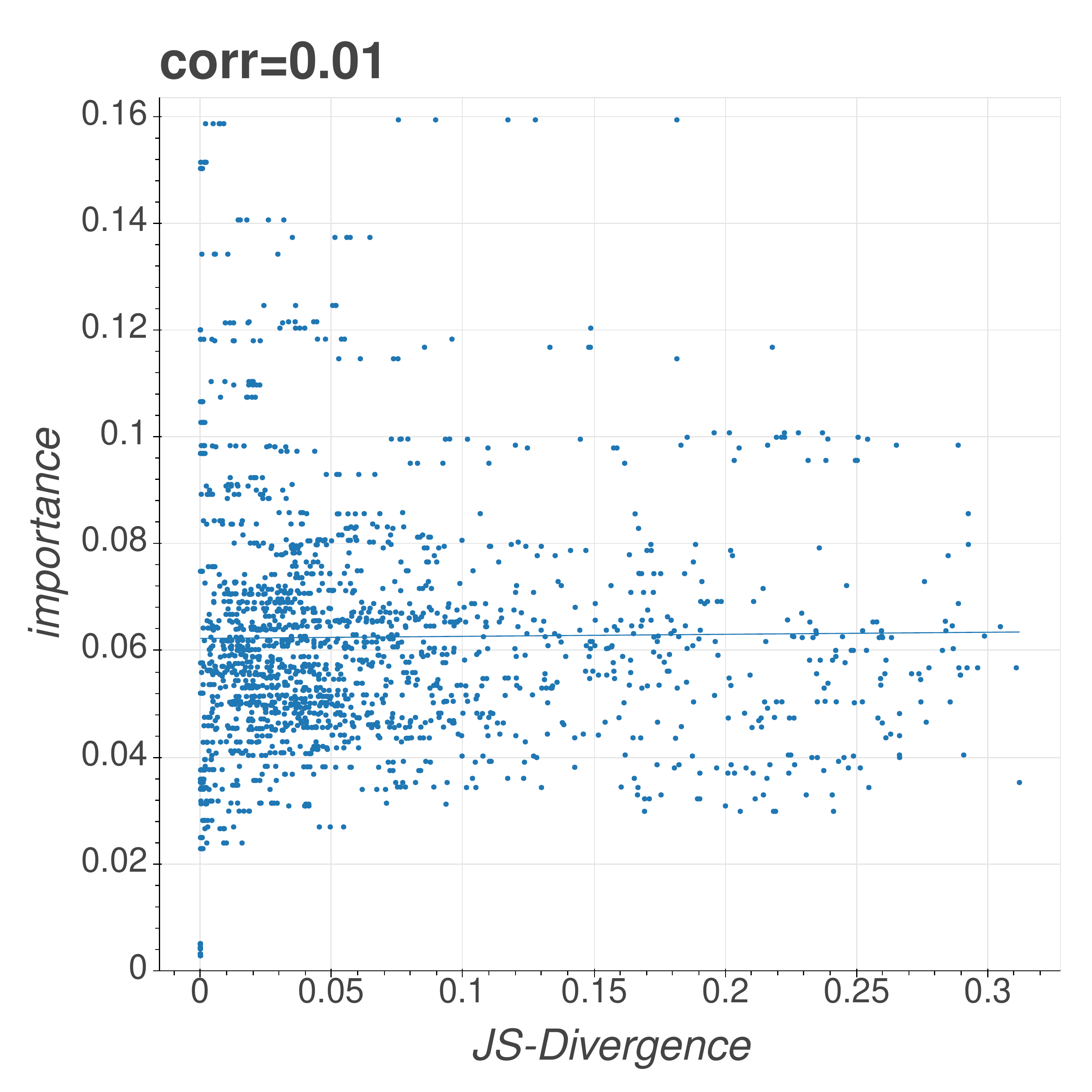}
    \includegraphics[width=0.24\linewidth]{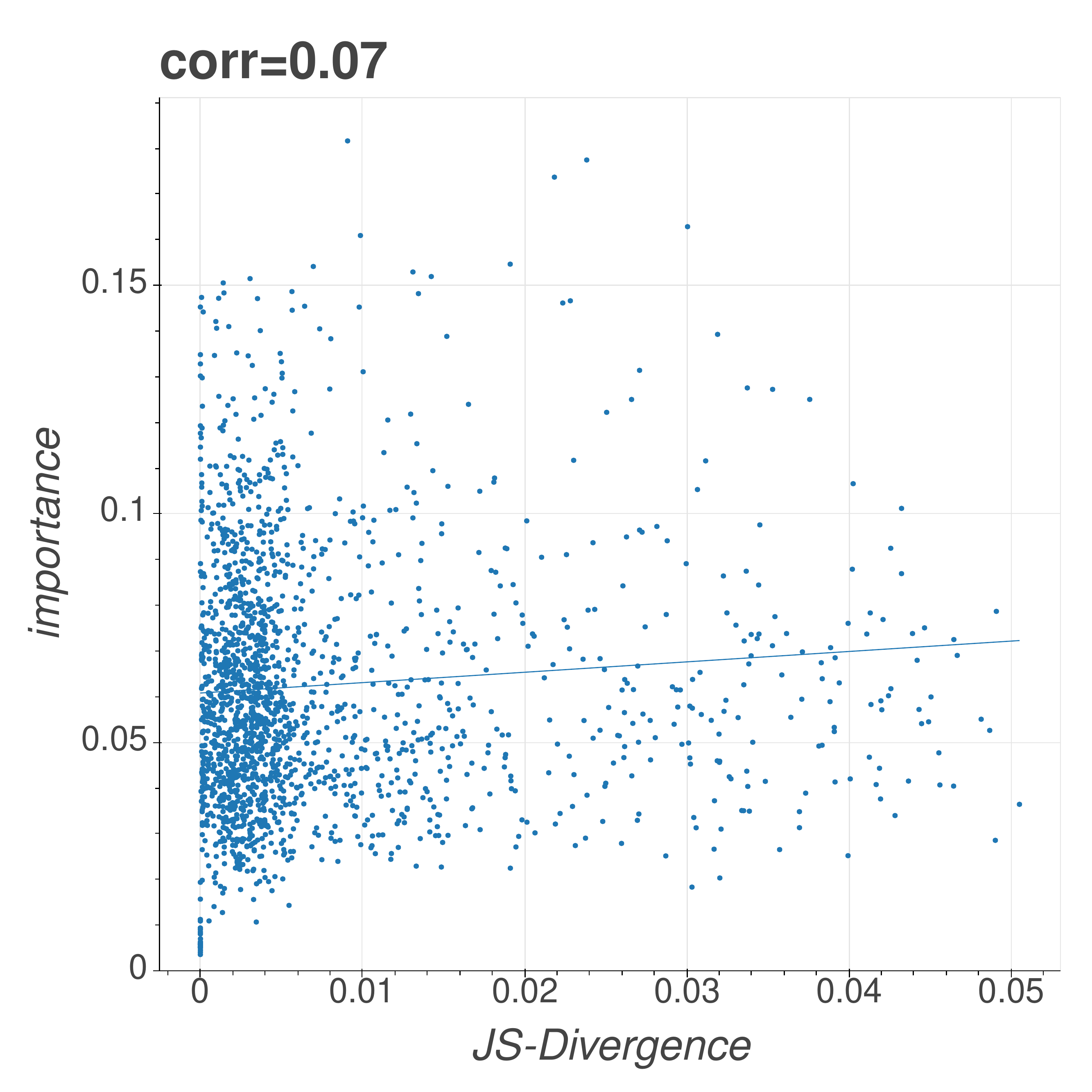}
    \includegraphics[width=0.24\linewidth]{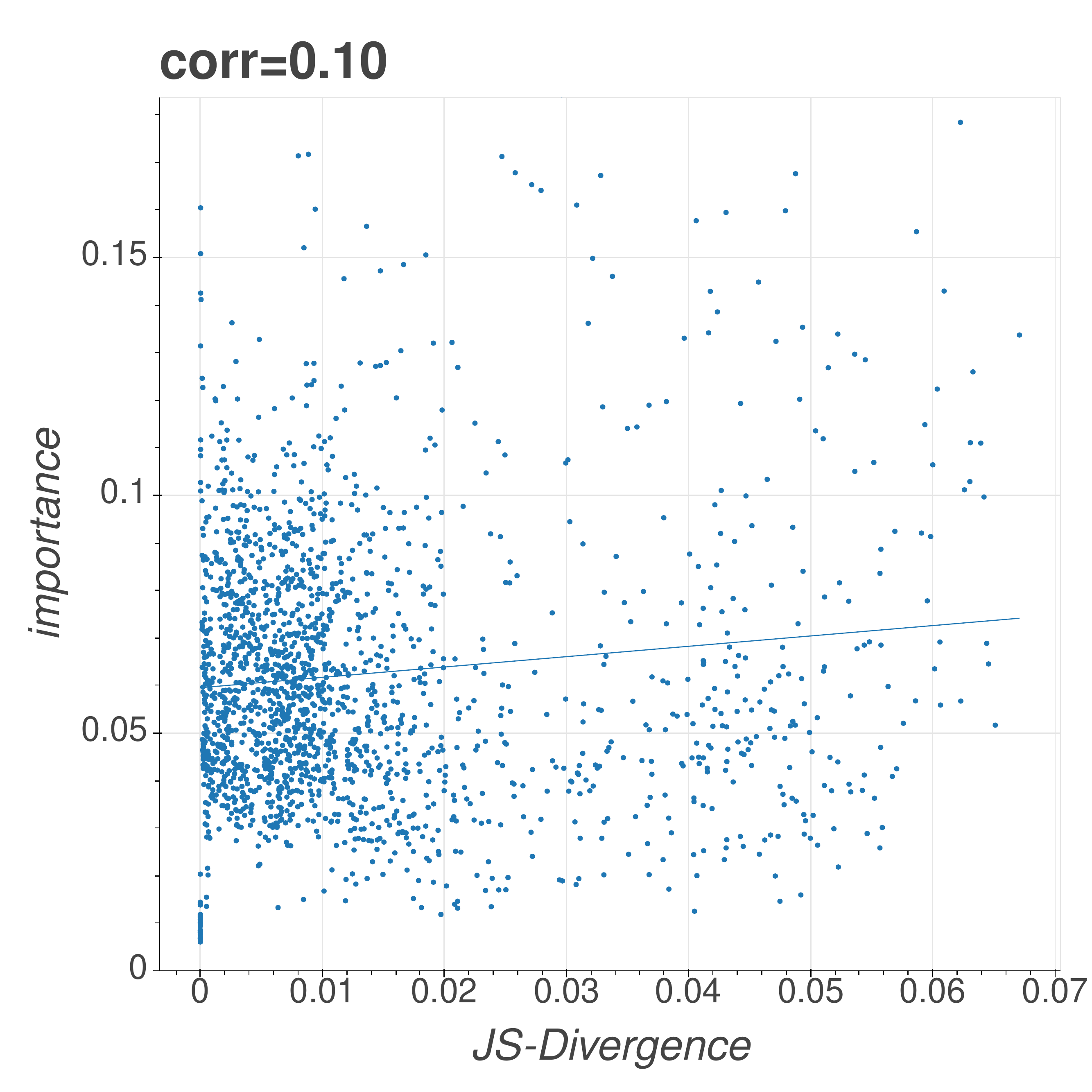}
    \includegraphics[width=0.24\linewidth]{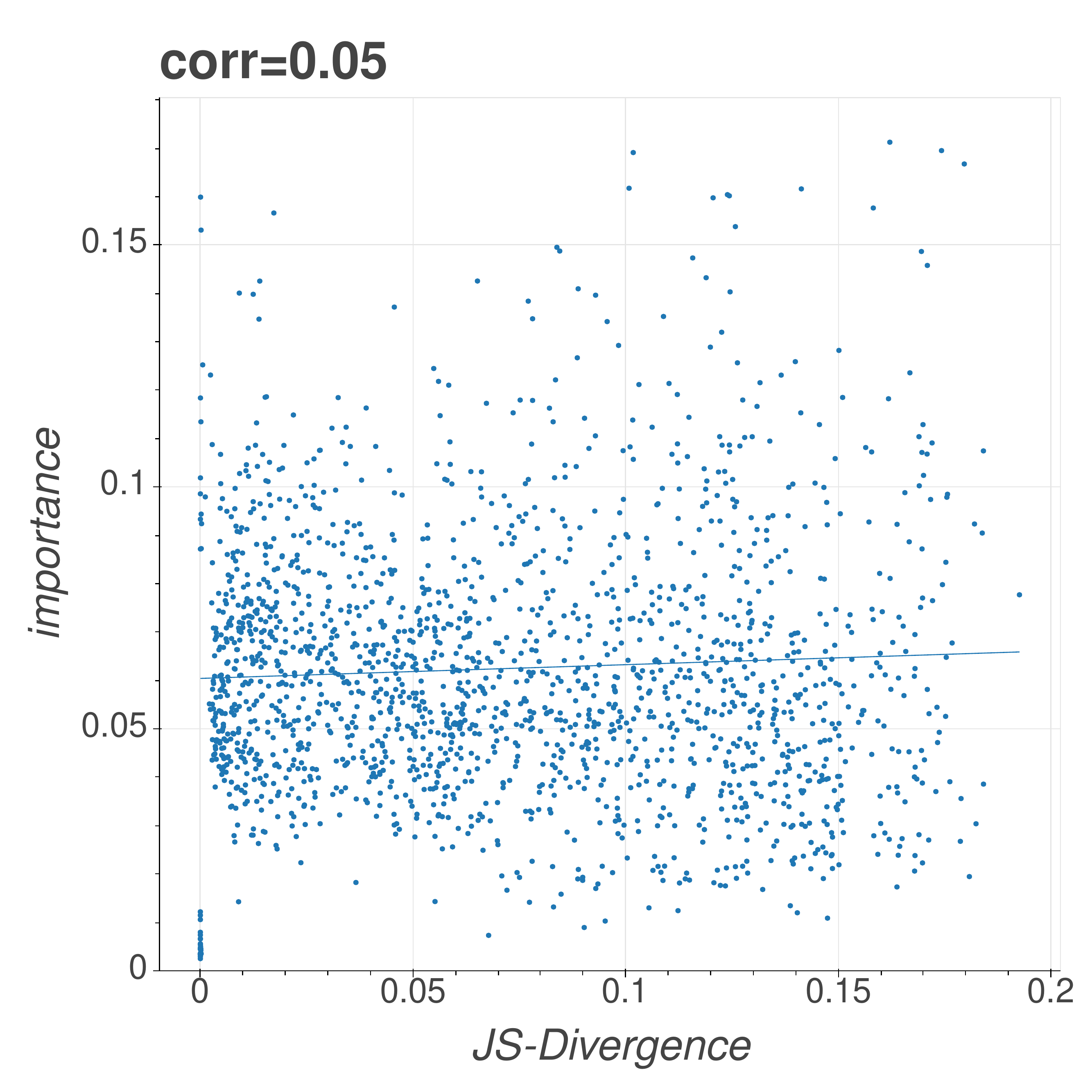}
    \includegraphics[width=0.24\linewidth]{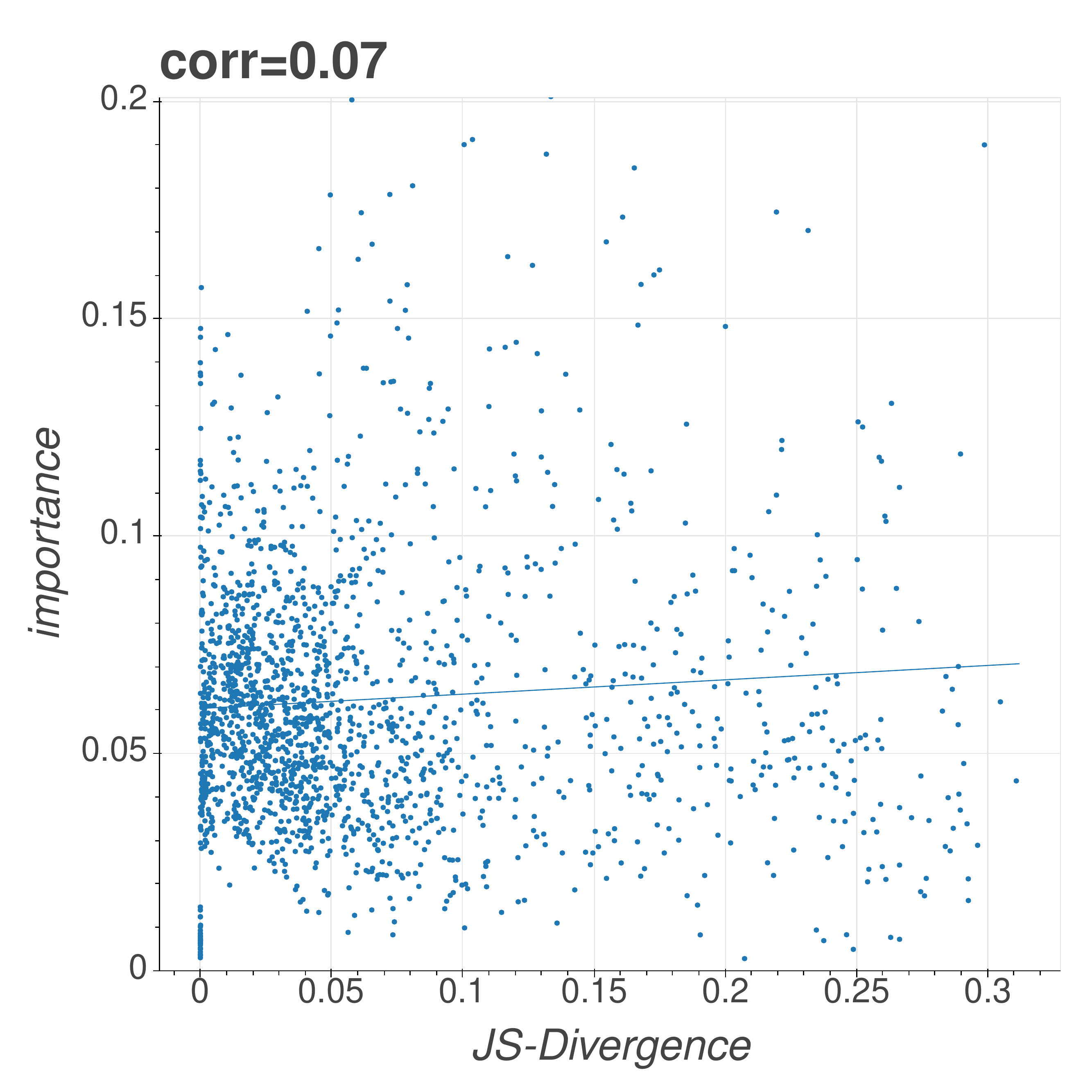}
    \caption{The relation between the JS-Divergence (x-axis) and the head importance (y-axis). Each point represents a head in a model. The solid lines represent linear regression. Figures in the upper row and the lower row plot the head importance estimated on the pre-train task and the downstream task respectively. The four columns are for the task CoLA, SST-2, QQP, and MNLI respectively.}
    \label{fig:hi-jsd-corr}
\end{figure*}

We plot the relation between head importance defined in \ref{eq:head-score-robeta} and JS-divergence discussed in section~\ref{sec:compare-behavior} in figure~\ref{fig:hi-jsd-corr}.

\end{document}